\documentclass[10pt,letterpaper]{article}
\usepackage[top=0.85in,left=2.75in,footskip=0.75in,marginparwidth=2in]{geometry}

\usepackage[utf8]{inputenc}

\usepackage{cite}

\usepackage{nameref,hyperref}

\usepackage{microtype}
\DisableLigatures[f]{encoding = *, family = * }

\raggedright
\usepackage{changepage}

\usepackage[aboveskip=1pt,labelfont=bf,labelsep=period,singlelinecheck=off]{caption}

\makeatletter
\renewcommand{\@biblabel}[1]{\quad#1.}
\makeatother

\usepackage{lastpage,fancyhdr,graphicx}
\usepackage{subcaption}
\usepackage{epstopdf}
\fancyheadoffset[L]{2.25in}
\fancyfootoffset[L]{2.25in}

\usepackage{color}

\definecolor{Gray}{gray}{.25}

\usepackage{graphicx}

\usepackage{sidecap}

\usepackage{wrapfig}
\usepackage[pscoord]{eso-pic}
\usepackage[fulladjust]{marginnote}
\usepackage{graphicx}
\usepackage{epstopdf}
\reversemarginpar

\begin{document}
\vspace*{0.35in}

\begin{flushleft}
{\Large
\textbf\newline{Motor Control Insights on Walking Planner and its Stability}
}
\newline
\\
Carlo Tiseo\textsuperscript{1,2,3,*},
Kalyana C Veluvolu \textsuperscript{4,*},
Wei Tech Ang\textsuperscript{1,2},
\\
\bigskip
\bf{1} Rehabilitation Research Institute of Singapore, 50 Nanyang Avenue N3-01a-01,Singapore 639798, Singapore;
\\
\bf{2} School of Mechanical \& Aerospace Engineering, Nanyang Technological University, 50 Nanyang Avenue N3-01a-01,Singapore 639798, Singapore;
\\
\bf{3} School of Informatics, University of Edinburgh, 1.17 Bayes Centre 47 Potterrow, EH8 9BT Edinburgh, UK;
\\
\bf{4} School of Electronics Engineering,Kyungpook National University, Daegu  702701, South Korea;
\\
\bigskip
* CT : ctiseo@ed.ac.uk ; KCV: veluvolu@ee.knu.ac.kr 

\end{flushleft}

\section*{Abstract}
The application of biomechanic and motor control models in the control of bidedal robots (humanoids, and exoskeletons) has revealed limitations of our understanding of human locomotion. A recently proposed model uses the potential energy for bipedal structures to model the bipedal dynamics, and it allows to predict the system dynamics from its kinematics. This work proposes a task-space planner for human-like straight locomotion that target application of in rehabilitation robotics and computational neuroscience. The proposed architecture is based on the potential energy model and employs locomotor strategies from human data as a reference for human behaviour. The model generates Centre of Mass (CoM) trajectories, foot swing trajectories and the Base of Support (BoS) over time. The data show that the proposed architecture can generate behaviour in line with human walking strategies for both the CoM and the foot swing. Despite the CoM vertical trajectory being not as smooth as a human trajectory, yet the proposed model significantly reduces the error in the estimation of the CoM vertical trajectory compared to the inverted pendulum models. The  proposed model is also able to asses the stability based on the body kinematics embedding in currently used in the clinical practice. However, the model also implies a shift in the interpretation of the spatiotemporal parameters of the gait, which are now determined by the conditions for the equilibrium and not \textit{vice versa}. In other words, locomotion is a dynamic reaching where the motor primitives are also determined by gravity.

\section{Introduction}

Owing to the development of bipedal robots and medical technologies for locomotor disabilities, there is an exponentially growing surge of interest in the bipedal equilibrium \cite{SAUNDERS1953, Kuo2007, Torricelli2016, McGeer1990, Carpentier2016,dobkin2017}. The earlier systematic investigation on locomotion strategies identified the pelvic rotation, pelvic tilt, knee stance flexion, lateral displacement of the pelvis, foot and knee mechanisms as parameters (gait determinants) that characterise the human locomotion and differentiate pathological behaviours \cite{SAUNDERS1953}. Subsequently, the works focused on the analysis of the dynamics based on the inverted pendulum model for bipedal locomotion, while the gait determinants were gradually set aside due to their qualitative nature \cite{Torricelli2016, Kuo2007}, models derived from the inverted pendulum model are currently deployed for both the analysis of human behaviour and bipedal robots controllers \cite{Pratt2006,Torricelli2016,Kuo2007,Hof2005,Caron2016,Carpentier2016}. 

The inverted pendulum does not accurately model the dynamics for step-to-step transition \cite{Kuo2007}. Therefore, models like the Zero Moment Point (ZMP) and the extrapolated Centre of Mass (XCoM) are required for the extension of the inverted pendulum model to human-like locomotion (i.e., not constraint to the sagittal plane) by identifying the synchronization condition between the legs in the task-space \cite{Popovic2004, Torricelli2016, Kuo2007,Hof2007, Caron2016,englsberger2015three}. However, these models require local optimisation algorithms to plan the walking trajectory in the task space, making the process computationally expensive for highly redundant mechanisms  \cite{Perrin2012, Carpentier2016, Caron2016, Laumond2015}. 

\begin{figure*}[ht]
     \centering
     \begin{subfigure}[b]{0.60\textwidth}
         \centering
         \includegraphics[width=\textwidth]{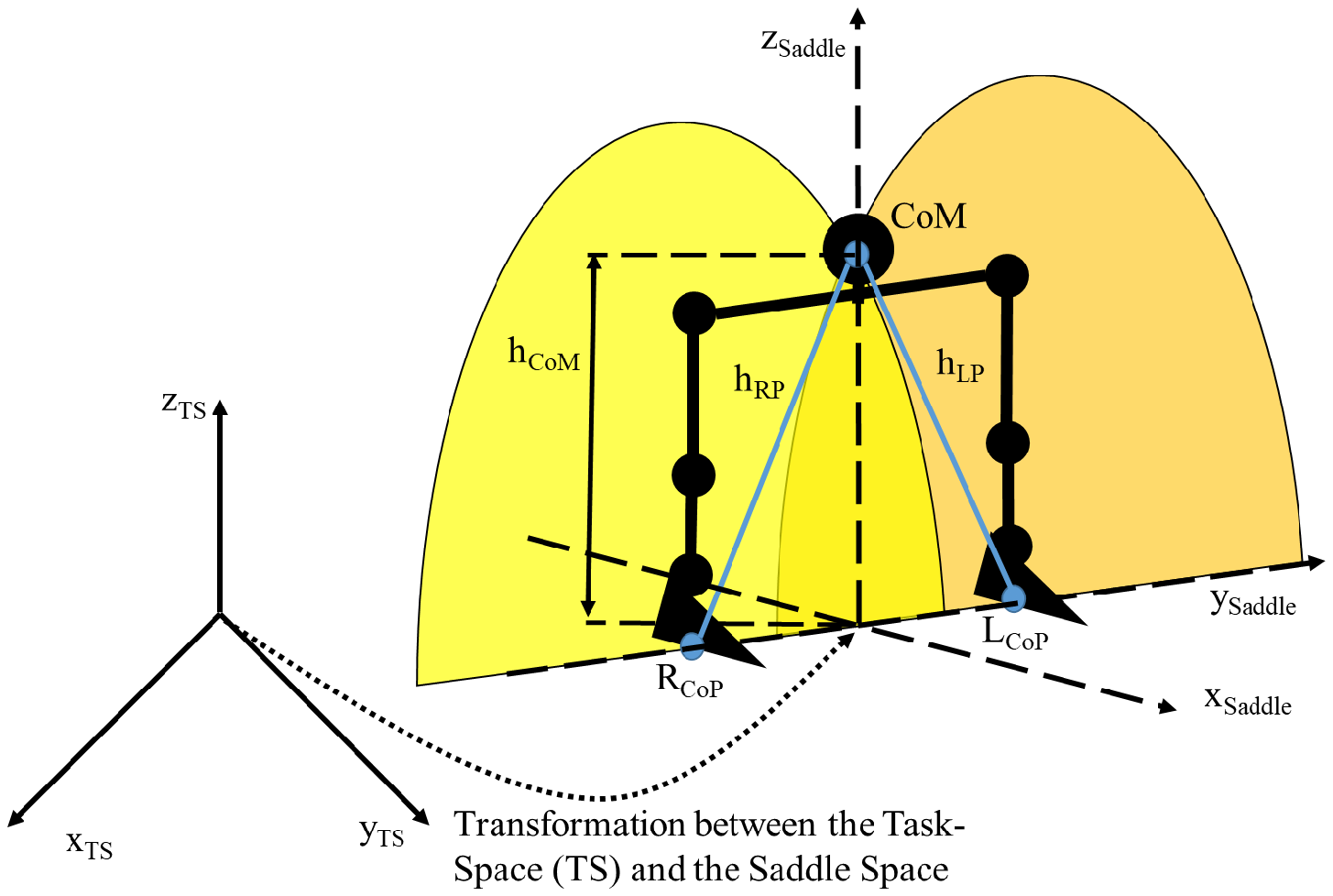}
         \caption{\centering\centering}
         \label{fig:01}
     \end{subfigure}
     \hfill
     \begin{subfigure}[b]{0.65\textwidth}
         \centering
         \includegraphics[width=\textwidth]{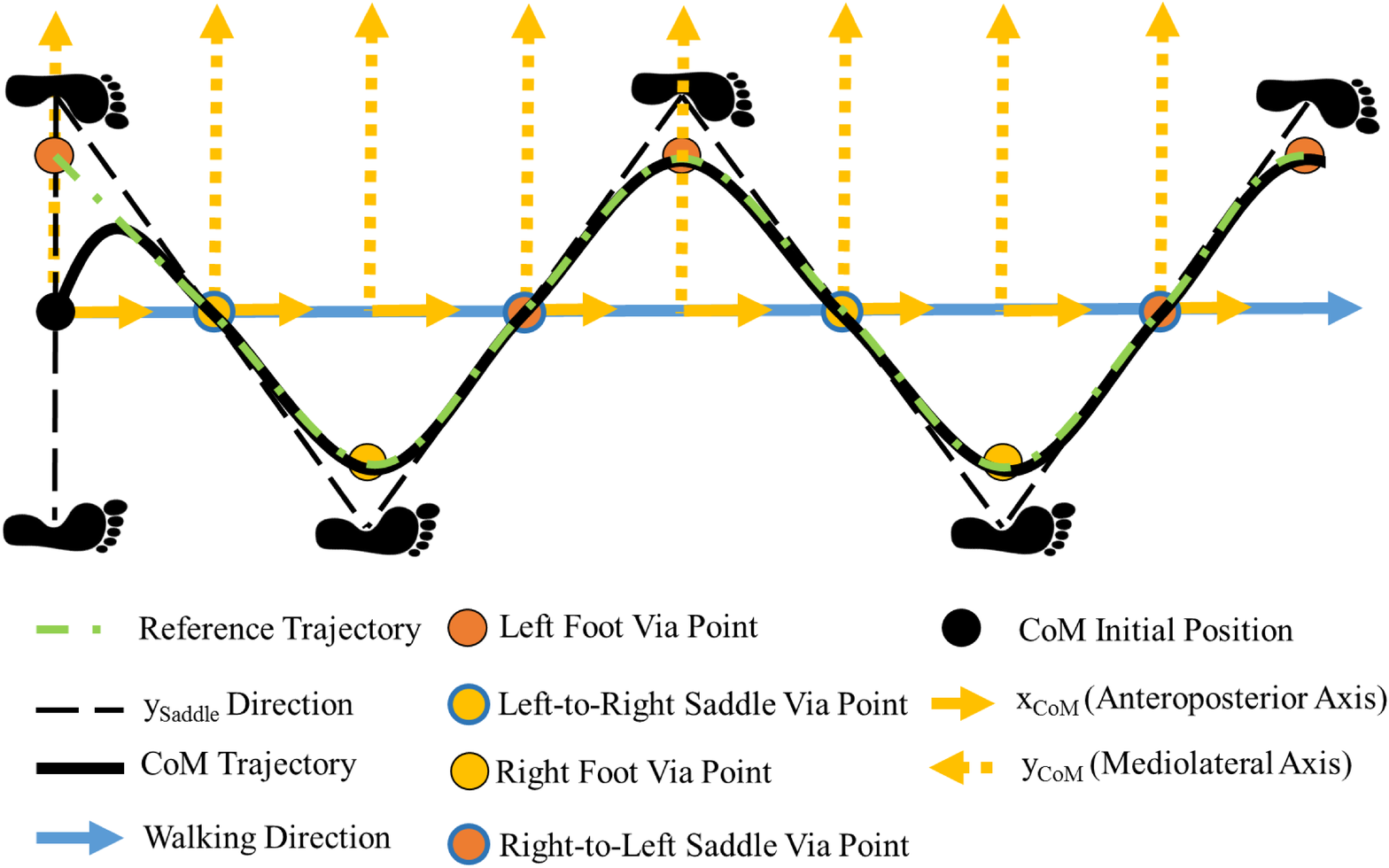}
         \caption{\centering}
         \label{fig:02}
     \end{subfigure}
    \begin{subfigure}[b]{0.65\textwidth}
         \centering
         \includegraphics[width=\textwidth]{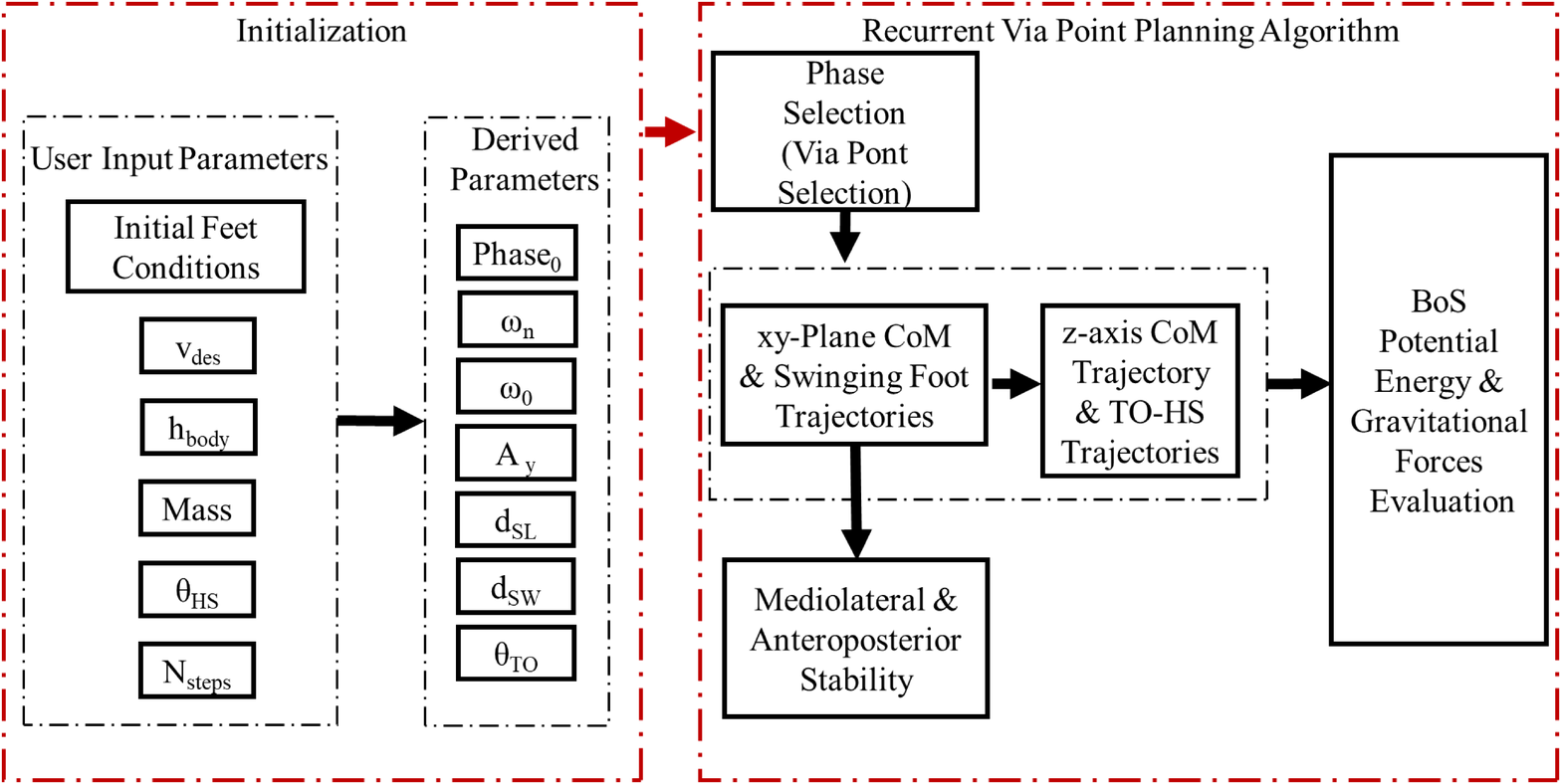}
         \caption{\centering}
         \label{fig:03}
     \end{subfigure}
        \caption{The Saddle Space and the generation of a recursive algorithm for bipedal locomotion planning. (a) The Saddle Space frame has the y-axis aligned with the principal direction connecting the two fulcra ($R_{CoP}$ and $L_{CoP}$) and the x-axis lying on the other saddle principal direction. (b) The proposed planner relies on the deployment of the potential energy fixed point to generate the desired trajectories for both the CoM and the swinging foot. This constrains the CoM trajectory to lie on $y_{Saddle}$ that nullifies the gravitational forces perpendicular to the CoM trajectory on the tangent plane \cite{Tiseo2016, Tiseo2018a}. (c) The proposed planner inputs are the initial feet position, the desired velocity, the body height, the mass, the desired HS angle when the foot hits the ground and the number of steps to be simulated. The planning algorithm derives all the other parameters from the inputs. For example, Phase$_0$ which is the initial gait phase, and the natural frequency of the inverted pendulum ($\omega_n$). Subsequently, the recurrent module starts and plan the selected number of steps. }
\end{figure*}

In this paper, we propose to apply a recently published analytical model of the potential energy generated by an anthropometric bipedal walker for the formulation of a planner in the task-space \cite{Tiseo2016, Tiseo2018a,Tiseo2018d}. The potential energy model allows to define a posture-dependent reference frame called Saddle Space that is aligned with the principal directions of the potential energy surface, as shown in Figure \ref{fig:01}. The y-axis of the Saddle Space ($y_{Saddle}$) is aligned with the principal direction between the three fixed points of the Saddle, and it includes the biped in the segment between the two feet where there is a stable dynamics \cite{Tiseo2018d}. On the other hand, the biped is always unstable along the x-axis, $x_{Saddle}$ \cite{Tiseo2016, Tiseo2018b, Tiseo2018c,Tiseo2018d}. 

The proposed task-space planner for straight walking can generate human-like trajectories, and it has been validated using the human motion capture data. This study aims to further the understanding of human locomotion motor control for the improvements of motor control models and rehabilitation therapies. Therefore, the choice to initially focus on a straight walking task has been based on the empirical evidence that humans have dedicated alignment strategies for locomotion, which are usually separated by a \textit{quasi}-straight walking path \cite{Sreenivasa2015}. The planner integrates commonly used parameters such as the XCoM and Base of Support (BoS) to evaluate stability from the locomotion kinematics. Furthermore, Lyapunov's stability analysis confirms that the BoS is a good cautionary estimation for the Region of Attraction. 

\section{Materials and Methods}
\label{Sec:2}

The planning algorithm is described in the first subsection, and includes both the generation of the Centre of Mass (CoM) and feet trajectories. It is observe from Figure \ref{fig:01} that the proposed method relies only on the Cartesian distances between the CoM and the two Centre of Pressures in the feet (COPs), which are the fulcra of the two pendula \cite{Tiseo2016,Tiseo2018d}. The methodology used in the evaluation of the planner performances is introduced at the end of the section.

\subsection {Straight Walking Planning Algorithm}
\label{Sec:2.1}
The transverse trajectory of the CoM is obtained by modelling its motion as a harmonic oscillator that moves at a constant speed. The planning algorithm divides a $2\pi$ cycle of the oscillator into four phases based on the 3 fixed points of the potential energy surface (2 maxima and 1 saddle point \cite{Tiseo2016,Tiseo2018d}). This cycle is exemplified in Figure \ref{fig:02}, and every phase is associated with a via point, as follows: 
\begin{enumerate}
	\item \textit{Left Foot Via Point }($L_{Fvp}$): It is defined as the desired maximum amplitude of the mediolateral trajectory at the chosen walking velocity during left support, which occurs when both feet are aligned on a segment perpendicular to the walking direction.
	\item \textit{Left-to-Right Saddle Via Point }($L_{Svp}$): It is defined as the desired saddle position associated with the chosen walking speed. The position of the saddle point is selected under the hypothesis that both pendula have the same maximum length; thus placing it in the middle of the segment connecting the two CoPs \cite{Tiseo2016,Tiseo2018d}.
	\item \textit{Right Foot Via Point }($R_{Fvp}$): It is the equivalent to $L_{Fvp}$ for the right support phase.
	\item \textit{Right-to-Left Saddle Via Point}($R_{Svp}$): It is defined as the position of the saddle during the transition from  right to left support.
\end{enumerate}
The recursive algorithm presented in Figure \ref{fig:03} has been derived from the equation to describe the CoM trajectory in the traverse plane in \cite{Tiseo2018d}. However, the architecture proposed requires also the identification of the required inputs and parameters, which can be obtained from the human motion capture data as explained later.

\subsubsection*{Marker set, CoM, CoP and BoS:}
The anthropometric parameters and the spatiotemporal parameters of the gait (e.g., the step length, $d_{SL}$, and step width, $d_{SW}$) are obtained from the KIT whole-body human motion databases (KITDB) \cite{Mandery2015}. The marker set and the method employed to calculate the CoM were derived from the four Iliac Markers. Hence, the CoM is placed in the middle of the segment connecting the frontal and the rear centres of the pelvis as defined by Iliac markers. Similarly, the CoPs are calculated as the mid-point along the segments joining the middle of the metatarsus (based on the two metatarsal markers) and the heel marker. It shall be noted that the CoP is defined as a geometrical point within the foot in our model, and it has been chosen based on its measure during standing reported in \cite{Hof2005,Tiseo2018c,Tiseo2018d}. The Base of Support (BoS) is introduced in our formulation in order to represent the range of motion of the CoP in the foot ($\simeq\pm 10 $ cm in the anteroposterior direction) \cite{Tiseo2018d}.

\subsubsection{CoM Trajectory Planning:}
\label{Sec:2.1.1}
The desired task-space CoM trajectory generation is based on the following equations \cite{Tiseo2016,Tiseo2018d}:
\begin{equation}
	\label{eq:01}
	\left\{
	\begin{array}{l}
		x_{CoMd}(t)=v_{des}t\\\\
		y_{CoMd}(t)=A_{y}\cos(\pi\omega_0 t +\phi)\\\\
		z_{CoMd}(t)=
		\left\{
		\begin{array}{lrcc}
			h_{CoM_L}(t), & if& Left & Support\\\\
			h_{CoM_R}(t), & if& Right & Support
		\end{array}\right.\\\\
	\end{array}\right.
\end{equation}
where $v_{des}$ is the desired walking speed, t is the time, and the other parameters are:
\begin{enumerate}
	\item $\omega_0=v_{des}/d_{SL}$ is the step cadence, where $d_{SL}$ is the step length.
	\item $\phi$ is the gait phase at t=0, and it is derived as follows
	\begin{enumerate}
		\item $\phi=0$: The CoM moves from $L_{Fvp}$ to $L_{Svp}$.
		\item $\phi=\pi/2$: The CoM moves from $L_{Svp}$ to $R_{Fvp}$.
		\item $\phi=\pi$: The CoM moves from $R_{Fvp}$ to $R_{Svp}$.
		\item $\phi=3\pi/2$: The CoM moves from $R_{Svp}$ to $L_{Fvp}$ .
	\end{enumerate}	
	\item $A_{y}=d_{SW}/(2\pi\omega_{0} d_{SL})$ is the mediolateral amplitude of the CoM movement, where $d_{SW}$ is the step width. This formulation for $A_y$ is obtained by imposing the trajectory of the CoM tangent to $y_{Saddle}$ during the step-to-step transition to maximise the stability \cite{Tiseo2018a,Tiseo2018b, Tiseo2018c}. This condition guides the CoM lateral trajectory along the segment connecting the two feet while passing for the saddle point during the double support.
	\item $z_{CoM_{d}}(t)$ depends on both the support foot of the gait phase and the length of the pendulum generated by the leg:
	\begin{enumerate}
		\item $h_{CoM_L}(t)=(h_{LP}(t)^2-(x_{CoMd}(t)-x_{LCoP}(t))^2-(y_{CoMd}(t)-y_{LCoP}(t))^2)^{0.5}$ where $x_{LCoP}(t)$ and $y_{LCoP}(t)$ are the coordinates of $L_{CoP}$.
		\item $h_{CoM_R}(t)=(h_{RP}(t)^2-(x_{CoMd}(t)-x_{RCoP}(t))^2-(y_{CoMd}(t)-y_{RCoP}(t))^2)^{0.5}$ where $x_{RCoP}(t)$ and $y_{RCoP}(t)$ are the coordinates of $R_{CoP}$.
	\end{enumerate}
	where both leg lengths, $h_{LP}(t)$ and $h_{RP}(t)$, depend on the ankle strategies.
\end{enumerate}	
Lastly, the anteroposterior trajectory of the foot is derived based on the hypothesis that the CoM stays on $y_{Saddle}$ also during the swing, where the bipedal structure has a stable dynamics (Figure \ref{fig:01}) \cite{Tiseo2016,Tiseo2018d}. Therefore, its trajectory is determined based on the condition that its CoP lies on the line connecting the CoM to the CoP that provides support \cite{Tiseo2016,Tiseo2018d}, which results in the following equation:
\begin{equation}
\left\{\begin{array}{lll}
x_{LCoP}(t)=
\frac{d_{SW}}{{m_{S_{//}}(t)}}+x_{RCoP}, & Right& Stance\\\\
x_{RCoP}(t)=\frac{d_{SW}}{{m_{S_{//}}(t)}}+x_{LCoP}, & Left& Stance
\end{array}\right.
\end{equation}
where  $m_{S_{//}}$ is the slope of $y_{Saddle}$.

\begin{figure*} [ht]
     \centering
     \begin{subfigure}[b]{0.49\textwidth}
         \centering
         \includegraphics[width=\textwidth]{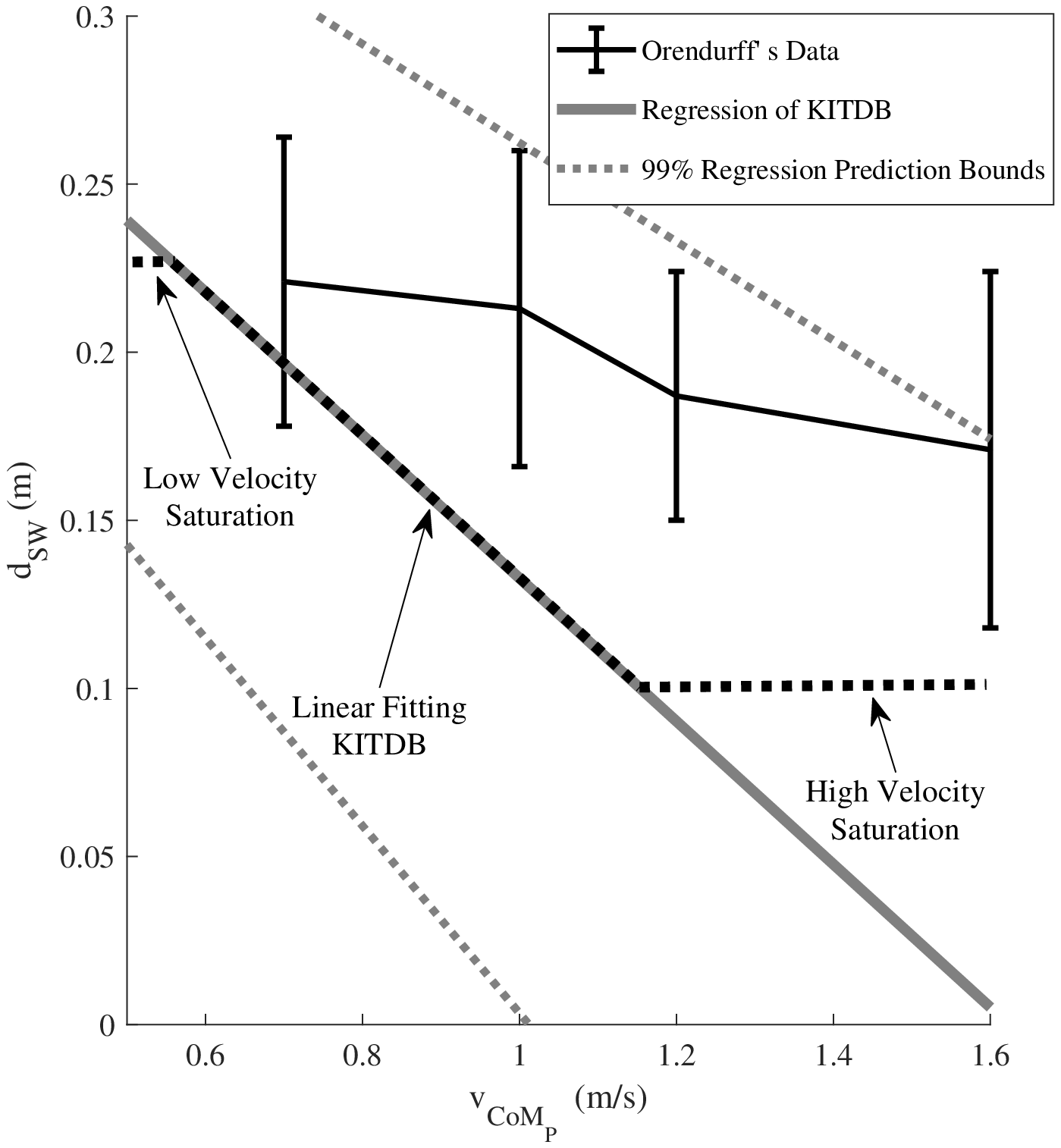}
         \caption{\centering}
         \label{fig:04}
     \end{subfigure}
     \hfill
     \begin{subfigure}[b]{0.49\textwidth}
         \centering
         \includegraphics[width=\textwidth]{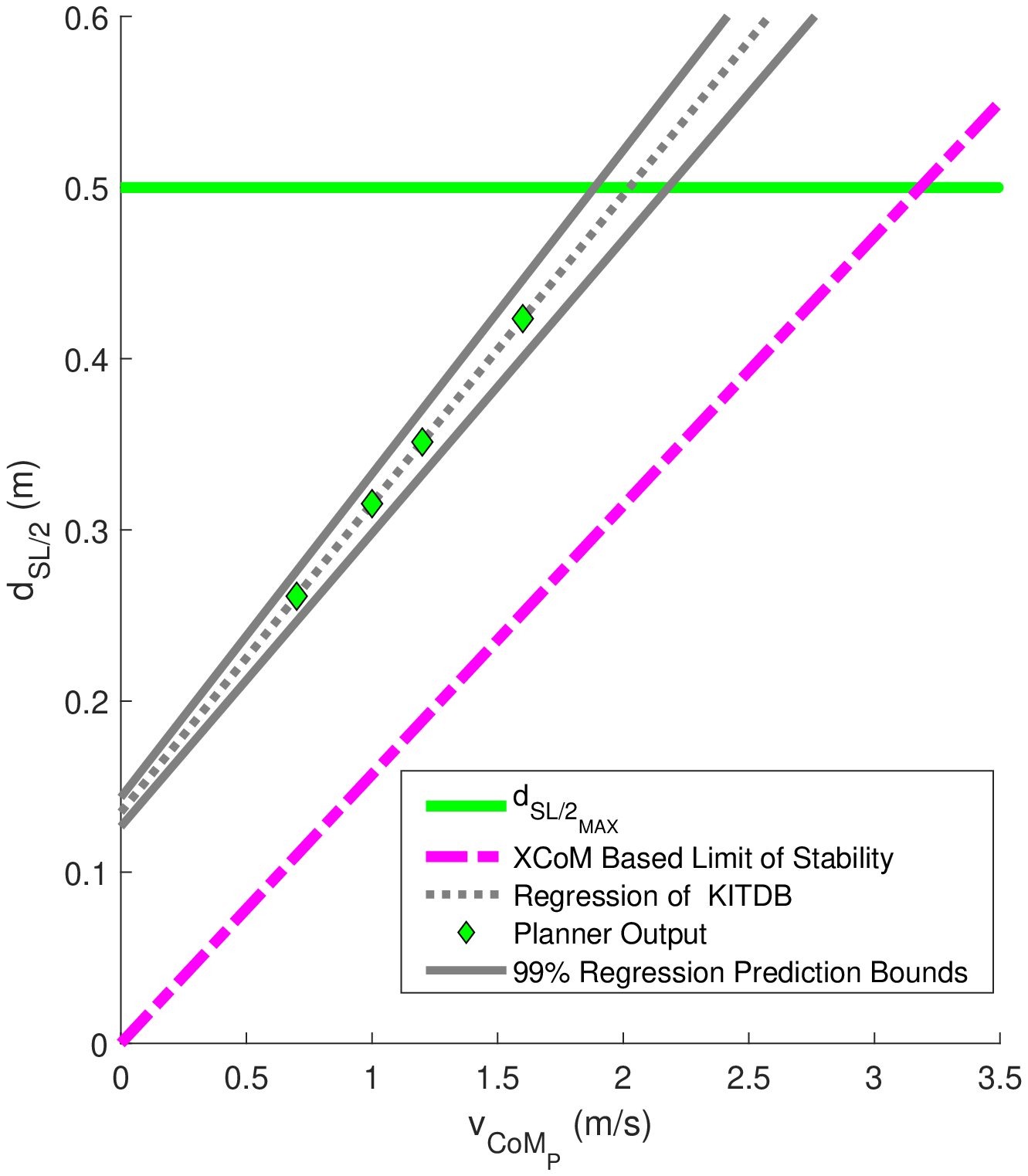}
         \caption{\centering}
         \label{fig:05}
     \end{subfigure}
    \caption{(a) The Step Width ($d_{SW}$) model is derived from the regression of motion capture data (MoCap Data) from KIT Data Base, and saturation in both high and low walking speeds. (b) The region of admissible Step Length ($d_{SL}$) extends along the abscissa up to the XCoM limit of stability, and its border along the y-axis is determined by the maximum reachable distance $d_{SL/2_{MAX}}$. They intersect at the peak velocity of about 3 [m/s], which is consistent with the results reported by Patnaik \textit{et al}\cite{Patnaik2015}.}
\end{figure*}
\subsubsection{Regression of Step Length and Step Width from the KITDB:}
\label{Sec:2.1.2}
The analysis of the data from the KITDB, corroborated by the literature, shows a linear relationship between the walking speed and the step length ($d_{SL}$) \cite{Orendurff2004, Hof2005, Collins2013, Kim2015}. On the other hand, the step width ($d_{SW}$) shows a more complex highly variable behaviour, which has been connected to both the lateral stabilisation and the energy optimisation of the gait strategies \cite{Collins2013, Kim2015}. 

The data from the KITDB employed for this work includes 58 straight walking trajectories, which are collected from 6 different subjects  (4 males and 2 females) \cite{Mandery2015}. Their statistical distribution of age, mass and height are (mean $\pm$ std) $25\pm 2$ years, $63.3\pm10.3$ kg, and $1.79\pm0.10$ m.

Our model for the selection of the step width is shown in Figure \ref{fig:04}, and it is described by the following equations:
\begin{equation}
	\label{eq:03}
	d_{SW} (v_{des})=
	\left \{
	\begin{array}{c}
		0.22 \textit{ }(m), v_{des}<0.6 \textit{ }(m/s)\\
		-0.2128 v_{des}+0.3456 \textit{ }(m), \textit{ o/w}\\
		0.10 \textit{ } (m), v_{des} > 1.1 \textit{ } (m/s)\\
	\end{array}\right.
\end{equation}
where the $R^2$ of the KITDB data regression is 0.157. It is worth noticing that despite the linearity of the relationship between the mean step width and the velocity, the high variability of the step width in human data determines the low $R^2$ value.

The step length model expressed as half of the step length represents the distance travelled by the CoM in the anteroposterior direction in each of the four phases introduced above. The result of the linear regression used for the generation of the reference behaviour as shown in Figure \ref{fig:05} can be defined as:
\begin{equation}
	\label{eq:04}
	d_{SL/2} (v_{des})=\frac{d_{SL} (v_{des})}{2}=0.1802 v_{des}+0.1351 \textit{ } (m)
\end{equation}
where the $R^2$ of the KITDB data regression is 0.98.

\subsubsection{Ankle Strategies and CoM Vertical Trajectory Planning:}
\label{Sec:2.1.3}

The ankle strategies are widely regarded to have a fundamental role in both balance and gait efficiency \cite{Hof2007,Kim2015,MyungheeKim2013,Ahn2012,Farris2015,McIlroy2003,Torricelli2016,Maki2007}. Therefore, we have decided to investigate the hypothesis that the CoM vertical trajectory error in the inverted pendulum can be mitigated by accounting for the ankle postures in accordance with the following equation:
\begin{equation}
	\label{eq:05}
	\Delta h_{TO/HS}(t)=
	\left\{
	\begin{array}{ll}
		2d_h\sin(\theta_{TO}(t)/2),& \textit{ if Toe-Off}\\
		2d_h \sin(\theta_{HS}(t)/2),& \textit{ if Heel-Strike}
	\end{array}\right.
\end{equation}
where $d_h=0.1$ m is the distance of the CoPs from the anterior and the posterior centres of rotation of the foot during Toe-Off and Heel-Strike (HS), respectively \cite{Hof2005}. Moreover, $\theta_{TO}$ and $\theta_{HS}$ are the trajectories of the angles between the feet's soles and the ground during TO and HS respectively, and they are described at the end of this section. Hence, the length of the pendulum is modeled as follows: 

\begin{equation}
	\label{eq:06}
	\left\{
	\begin{array}{l}
		h_{RP}(t)=
		\left\{
		\begin{array}{ll}
			l_p+\Delta h_{R_{TO/HS}}(t),& \textit{ during TO/HS}\\
			l_p,& \textit{ otherwise}
		\end{array}\right.\\\\
		h_{LP}(t)=
		\left\{
		\begin{array}{ll}
			l_p+\Delta h_{L_{TO/HS}}(t),& \textit{ during TO/HS}\\
			l_p,& \textit{ otherwise}
		\end{array}\right.
	\end{array}\right.
\end{equation} 
where $l_p=0.57h_{body}$ is the length of the pendulum based on anthropometric measures \cite{Virmavirta2014}, and $\Delta h_{L_{TO/HS}}$ is defined in equation (\ref{eq:05}).

\begin{figure*}[ht]
     \begin{subfigure}[b]{0.6\textwidth}
         \centering
         \includegraphics[width=\textwidth]{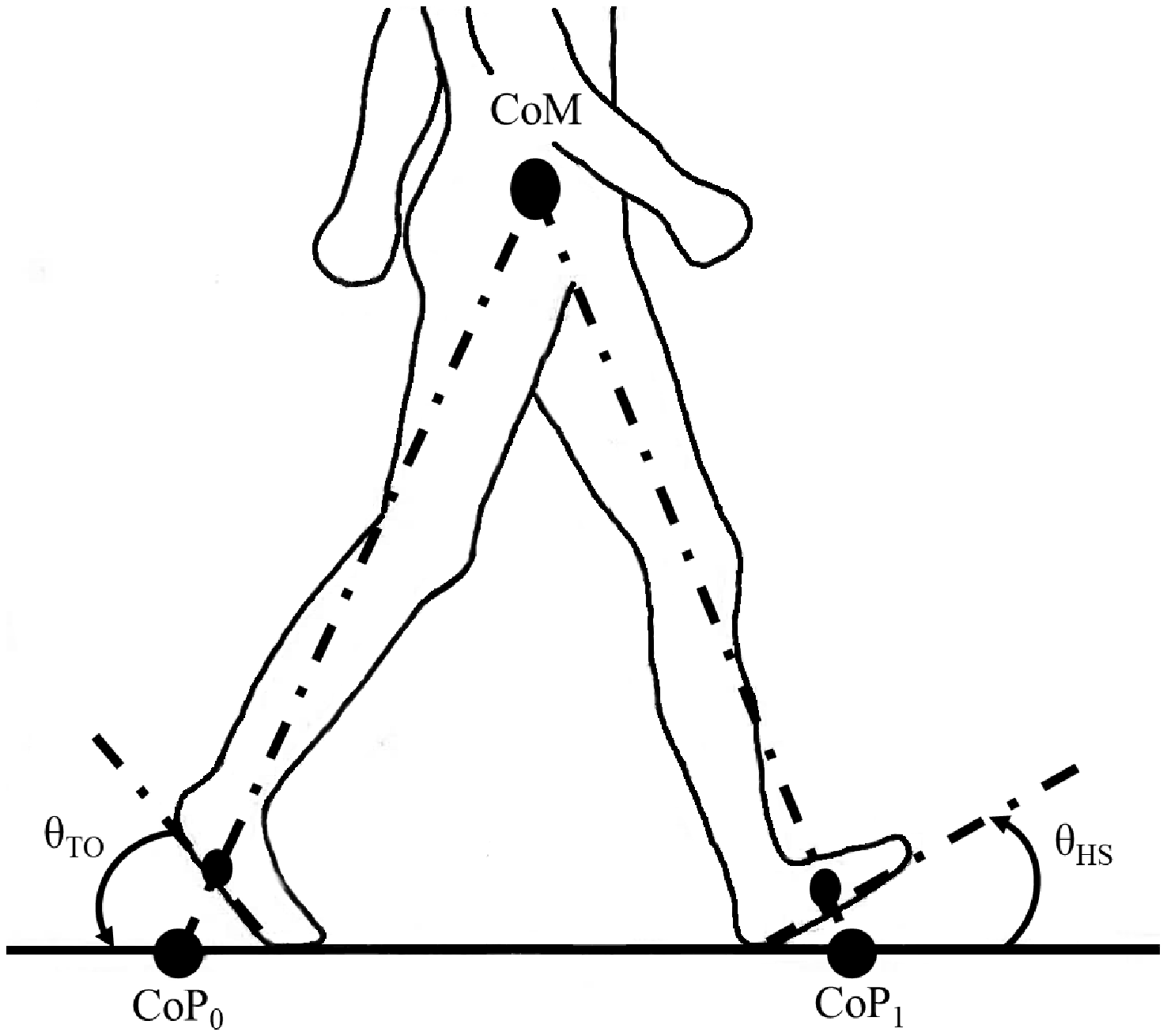}
         \caption{\centering\centering}
         \label{fig:06}
     \end{subfigure}
     \hfill
     \begin{subfigure}[b]{0.4\textwidth}
         \centering
         \includegraphics[width=\textwidth]{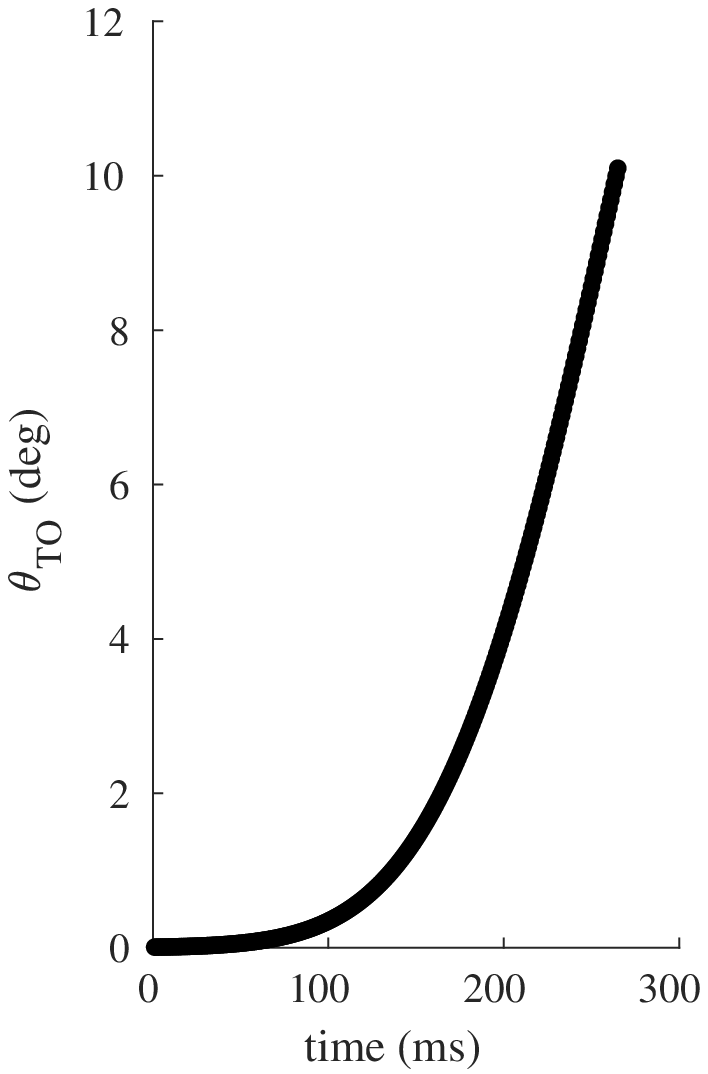}
         \caption{\centering}
         \label{fig:12}
     \end{subfigure}
        \caption{(a) The elongations of the pendulum lengths required for reaching $CoP_0$ and $CoP_1$ are derived from the length of the circumference chord, which is centred in the pivoting points with the ground and has a radius $d_h=0.1$ m. The pivot has been placed in the heel during the HS, and it is located in the articulation between the metatarsus and the phalanges during TO. (b) Sample trajectory of the TO angle.}
\end{figure*}

The parameters required for the generation of the ankle strategies are calculated based on the desired amplitude of the CoM vertical trajectory at the chosen walking speed. This relationship between the vertical amplitude and velocity is obtained from the KITDB, as follows:
\begin{equation}
	\label{eq:07}
	\Delta Z_{CoM}(v_{des})=0.02656 v_{des} +0.002575 \textit{ }(m) 
\end{equation}
where $R^2$ of the KITDB data regression is 0.2015. It is worth noticing that the relationship between the ankle angle and the velocity is characterised by the variability of the human data as evident from its low $R^2$ value. Furthermore, the behaviour observed is similar to the one observed in \cite{Orendurff2004}. Hence, equation (\ref{eq:07}) is used to determine the following parameters required to calculate the $\theta_{HS}(t)$ and $\theta_{TO}(t)$.

\begin{equation}
	\label{eq:08}
	\left\{
	\begin{array}{ll}
		t_{HS}&=\frac{1}{2\omega_0}-\frac{d_h(1-cos(\theta_{HS}))}{v_{des}}+t_0 \\\\
		l_{p0}&=((l_p-\Delta Z_{CoM}(v_{des}))^2+\\&+(x_{CoMd}(t_{HS})-x_{CoP0})^2\\&+
		(y_{CoMd}(t_{HS})-y_{CoP0})^2)^{0.5}\\\\
		Max(\theta_{TO})&=2\arcsin(\frac{l_{p0}}{2d_h})
	\end{array}\right. 
\end{equation}
where $Max(\theta_{HS})$ is the Heel Strike angle provided as an input, $t_0$ is the starting time of the current phase, and $x_{CoP0}$ and $y_{CoP0}$ are the coordinates of the CoP before the starting of the TO, as shown in Figure \ref{fig:06}. 

The trajectories of the ankle angles have been modelled for both TO and HS strategies with the error function (erf) available in Matlab (Mathworks Inc), as follows:
\begin{equation}
	\label{eq:A1}
	\left\{\begin{array}{ll}
		\theta_{HS}(t)&=-Max(\theta _{HS})\times erf(\frac{(t-t_{HS})}{0.11}+\\&+Max(\theta _{HS})\\\\
		\theta_{TO}(t)&=Max(\theta _{TO})\times erf(\frac{(t-t_{HS})}{0.11})+\\&+Max(\theta _{TO})
	\end{array}\right.
\end{equation}
where $t_{HS}$ is the instant where the sigmoid curve is centred at the HS,  $Max(\theta _{TO})$ is the TO angle when the HS occurs and $Max(\theta _{HS})$ is the desired HS angle that is provided as the input. A sample trajectory for the TO strategy is shown in Figure \ref{fig:12}.

\subsubsection{BoS Geometry}
The analysis of the system stability is based on the BoS definition that guarantees the necessary condition for stability and its borders are the Margins of Stability (MoS) \cite{Tiseo2016,Tiseo2018d}. In other words, it describes the regions where the system can be stabilised in the absence of external perturbations. The BoS model used in this work has been proposed in \cite{Tiseo2016, Tiseo2018d} and validated data in \cite{Tiseo2018a}. The BoS model in saddle space coordinates:
\begin{equation}
\label{Eq:extra}
BoS \le
\left\{ \begin{array}{lllc}
y_{S}^2+x_{S}^2=( Y_{sLF})^2,& if&  y_{S}\ge 0\\
& \& & x_{S}\le+d_h\\\\
y_{S}^2+x_{S}^2=(Y_{sRF})^2,& if& y_{S}\le 0\\
& \& & x_{S}\ge-d_h\\\\
\textit{Otherwise:}& &\\\\
x_{S}= +d_h,&  if & x_{S} \ge 0 & \\\\
x_{S}= -d_h,&  if & x_{S} \le 0 &\\\\
\end{array}\right.
\end{equation}
where ($x_{S}$, $y_{S}$) are the coordinates in the saddle space, $Y_{sLF}$ and $Y_{sRF}$ are the coordinates of the left and right CoP in the Saddle Space. The BoS can then be projected in the TS using the following relationship:

\begin{equation}
\label{eq:extra2}
\begin{array}{ll}
\left[
\begin{array}{l}x_{TS}\\
y_{TS}\end{array}\right]&=R(\lambda)\vec{x}_{S}+\vec{x}_{S0}=\\
&=\left[\begin{array}{cc} cos(\lambda) & -sin(\lambda)\\
sin(\lambda) &  cos(\lambda)\end{array}\right]
\left[\begin{array}{c}
x_{S}\\
y_{S}\end{array}\right]+\\\\&+\left[\begin{array}{c}
x_{S0}\\
y_{S0}\end{array}\right]
\end{array}
\end{equation}
where  $\lambda$ is the angle between $x_{TS}$-axis and the $x_{S}$-axis.
\subsection{Stability Supervision}
\label{Sec:2.2}
The stability analysis for bipeds has always been a challenging problem. Particularly, the unavailability of an appropriate dynamic model that can fully capture the complex human-like bipedal locomotion makes it difficult to define the general stability criteria, used in the trajectory planning \cite{Font-Llagunes2009,Manchester2010,Hof2008,Buschmann2015,Lei2006,Pratt2006,Torricelli2016,Kuo2007}. For example, the ZMP model evaluates the foot placement which is then used to compute the desired CoM trajectory with a stabilisable behaviour via optimisation algorithms \cite{Pratt2006, Perrin2012,Carpentier2016}. 
We proposed two kinematic based stability criteria in this work to evaluate if a movement is compatible with the stability. Hence, they can be used either for the definition of constraints for motion planning optimisation algorithms, or for the stability evaluation from the movement kinematics. 

\subsubsection{XCoM and Step Stability}
\label{Sec:2.2.1}
The step stability during walking is evaluated using the XCoM stability criterion and the maximum reachable distance. In fact, the XCoM allows to identify the minimum step length required for walking at a certain speed \cite{Hof2005,Hof2007,Hof2008}. Therefore, it helps to define the criterion for the step stability as shown in Figure \ref{fig:05}.
\begin{equation}
	\label{eq:09}
	d_{SL/2} \textit{ }\in \textit{ } (d_{SL/2_{min}}=\frac{v_des}{2w_n} \textit{ },\textit{ } d_{SL/2_{Max}} )
\end{equation}
where $w_n$ is the natural frequency of the inverted pendulum; $d_{SL/2_{min}}$ is half of the minimum step defined by the natural frequency of the pendulum; $d_{SL/2_{Max}}$ is half of the maximum step that can be performed by the biped. This allows us to establish the following metrics for the step stability:
\begin{equation}
	\label{eq:10}
	\left \{
	\begin{array}{l}
		S_{SL_{Pend}}(t,v_{CoMp})=\frac{M_{d_{SL/2}}(t)-d_{SL/2_{min}}(v_{CoMp})}{d_{SL/2}(v_{CoMp})-d_{SL/2_{min}}(v_{CoMp})} \\\\
		S_{SL_{Jump}}(t,v_{CoMp})=\frac{M_{d_{SL/2}}(t)-d_{SL/2_{Max}}(v_{CoMp})}{d_{SL/2}(v_{CoMp})-d_{SL/2_{Max}}(v_{CoMp})}
	\end{array}\right.
\end{equation}
where $v_{CoMp}$ is the walking velocity, $M_{d_{SL/2}}$ is the half-step length expected as from the biped and $d_{SL/2}$ is the desired behaviour. In summary, equation (\ref{eq:09}) describes how to define the stable step length strategy for a generic bipedal structure. On the other hand, equation (\ref{eq:10}) evaluates the selected strategy against an optimal strategy thus providing a quantitative evaluation of the distance from the margins of stability. 
\subsubsection{Mediolateral Stability}
\label{Sec:2.2.2}
The ML stability evaluation is commonly based on the criterion that the CoM has to be constrained between the two CoPs in the ML direction \cite{Vlutters2016,VanMeulen2016,McAndrewYoung2012,Lugade2011,Hof2007}.
\begin{equation}
	\label{eq:11}
	S_{SW}(t)=1-\frac{2|y_{CoM}(t)|}{d_{SW}(v_{CoMp})}
\end{equation}

\subsection{Stability Analysis: Relationship between BoS and Dynamic Stability}
\label{Sec:2.3}
To study the stability in the general case, let us consider the system energy described by the following equation:
\begin{equation}
	\label{eq:12}
	E=U+K+W+E_P
\end{equation}
where $U$ is the potential energy of the CoM, $K$ is the kinetic energy, $W$ is the active work, and $E_p$ are the external perturbations. Therefore, the condition for achieving stability in a desired posture is:
\begin{equation}
	\label{eq:13}
	Max(W)\ge U_{MoS}-U_{CoM}-K-E_p
\end{equation}
where $U_{CoM}$ and $U_{MoS}$ are the potential energies at the current CoM position and expected intersection with MoS, respectively.  If the condition in equation (\ref{eq:13}) cannot be satisfied, then the system cannot reach a static equilibrium in the existing feet posture, and it should be reconfigured. In other words, a biped is stable as long as it is able to actively adsorb the excess energy (equation (\ref{eq:13})) to stop in the current feet posture, or if it can implement a stable locomotor strategy as described in equations (\ref{eq:10}) and (\ref{eq:11}). In conclusion, equation (\ref{eq:13}) enables to compute the region of attraction that the bipeds can generate around its CoM for a given posture. Further, it serves as a tool for the analysis of movement stability that can be represented as a continuous transition between stable postures.  
\subsubsection{BoS as Region of Attraction}
\label{Sec:2.3.1}
The region of attraction can be defined as the set of points where the system is Lyapunov's stable \cite{Manchester2010, Majumdar2013}. If we choose equation (\ref{eq:12}) as Lyapunov' s candidate, then the system is stable if and only if the following condition is satisfied:
\begin{equation}
	\label{eq:15}
	\dot{E} = \dot{U}+ \dot{K}+ \dot{W}+ \dot{E}_P \le 0
\end{equation}
in the absence of the external perturbations. The condition for Lyapunov's stability of a trajectory (\textit{C}) between the points A and B is:
\begin{equation}
	\label{eq:16}
	W_\textit{C} \ge U(B) -U(A)+K -K_{des}
\end{equation}
where $W_{\textit{C}}$ is the maximum energy that the system can actively dissipate along \textit{C}, and $K_{des}$ is the kinetic energy of the desired trajectory.

\subsubsection{BoS as Regions of Finite Time Invariance}
\label{Sec:2.3.2}
The region of attraction theory is valid for time-invariant systems, while our model is time variant because the potential energy depends on the body configuration. However, if the system is considered as time-invariant for small time intervals, then this set of points can be defined as a Region of Finite-Time Invariance \cite{Majumdar2013}. This method is commonly used for the control of systems in highly unstructured environments, which can be accurately predicted for short periods of time \cite{Majumdar2013}. Thus, using the BoS as the set of reachable stable points at a given configuration, the gravitational forces can be regarded as time-invariant within the BoS at every instant. Consequently, the BoS considered as a Region of Finite-Time Invariance, and the instantaneous stability can be evaluated using equation (\ref{eq:16}). 

\subsubsection{E-BoS for Global Stability and I-BoS for Local Stability:}
\label{Sec:2.3.3}
The regions of attraction and finite time invariance allows to evaluate the stability of the system used in different scenarios. 

The region of attraction is more suited for analysing the global stability of a task because it allows the estimation of the margin of stability for the chosen end-posture. Hence, it provides the expected region of attraction (Expected BoS, E-BoS) for a future posture. Instead, the control of the trajectory requires to consider the local stability conditions that take into account both the local dynamics and the unforeseeable perturbation. Therefore the Region of Finite time-invariance (Instantaneous BoS, I-BoS) is required for carrying out such evaluation.

An example of the locomotion-related scenarios that allows understanding the difference between E-BoS and I-BoS is:
\begin{itemize}
	\item \textit{Unperturbed walking on a flat surface}: This being a deterministic walking condition, there are no unpredictable external factors; hence, the E-BoS is a sufficient condition for stability.
	
	\item \textit{Perturbed gait}: People walking in everyday living environments are subject to a multitude of perturbations that cannot be predicted during planning. If we consider an unexpected push from behind, it introduces an unforeseeable increase in the kinetic energy along the forward direction. Hence, the I-BoS provides local information required for equation (\ref{eq:16}) to evaluate the system stability and to plan a response strategy in the altered state. 
\end{itemize} 

\subsection{Simulations}
The simulations have been conducted with Matlab 2016 (Mathworks inc., USA) running on a Lenovo Y50 equipped with an Intel i7-4700HQ and 16 GB of memory. The simulation time is calculated with the \textit{run and time} function included in the software. The time step used in the simulations is 80 ms, which is based on the fastest Central Nervous System (CNS) response time to balance the perturbations \cite{Maki2007,McIlroy2003}.

The simulations have been performed at velocities of 0.7, 1.0, 1.2 and 1.6 (m/s). The HS angles used are 5, 10 and 15 (deg). The feet postures have been selected to have the initial gait phase ($Phase_0$) equal to 0. The mean values of the KIT data are used for the body height and the mass parameters. The simulations have been executed for 2 steps or 1 stride, which leads to a change of gait phase from 0 to $\pi$, as shown in Figure \ref{eq:10}. An additional set of simulations has been performed with the Matlab timing function to evaluate the planning time for 1 and 10 consecutive strides, which includes the stability evaluation with equations (\ref{eq:10}) and \ref{eq:11}. Furthermore, the planner also calculates the gravitational forces and the MoS calculated with the model presented in \cite{Tiseo2016,Tiseo2018d}.

\begin{figure*}[ht]
     \centering
     \begin{subfigure}[b]{0.49\textwidth}
         \includegraphics[width=\textwidth]{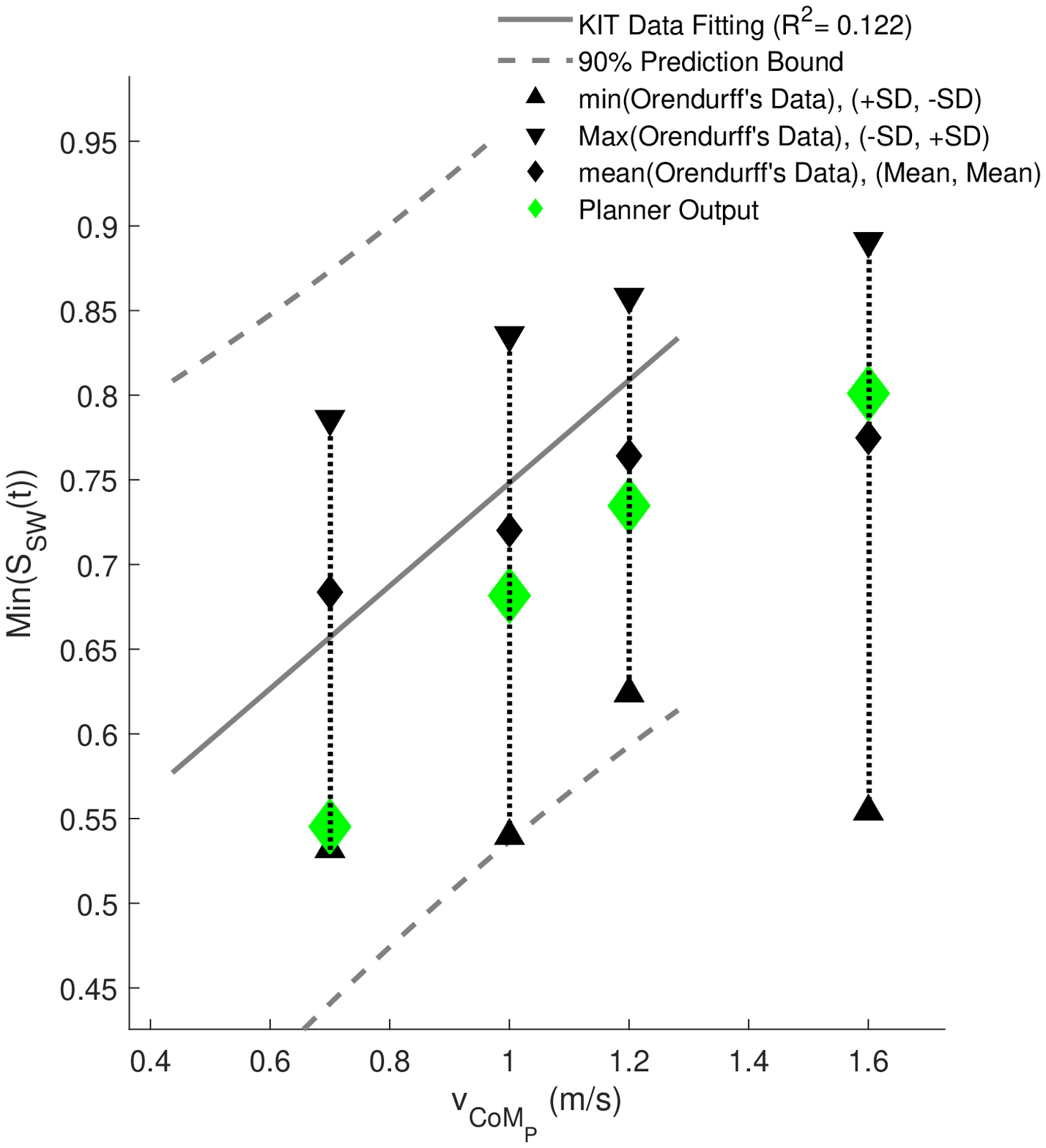}
         \caption{\centering\centering}
         \label{fig:07}
     \end{subfigure}
     \begin{subfigure}[b]{0.49\textwidth}
         \includegraphics[width=\textwidth]{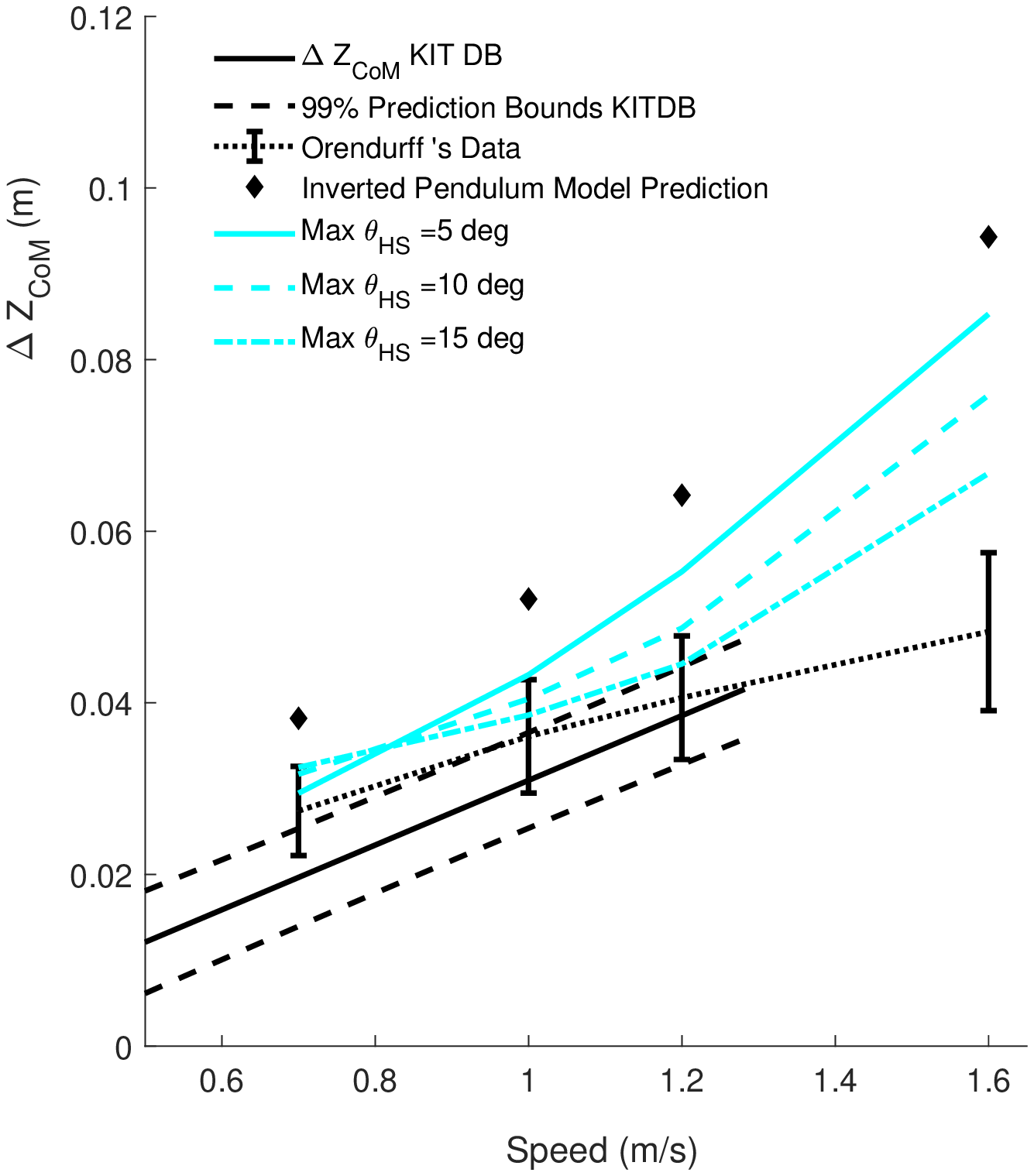}
         \caption{\centering}
         \label{fig:08}
     \end{subfigure}
        \caption{(a) The evaluation of the planner's performance along the mediolateral direction has been conducted by evaluating the minimum of the lateral stability. (b) The introduction of the ankle strategies can justify the discrepancy in the vertical CoM trajectories between the human behaviour and the inverted pendulum model.}
\end{figure*}
\section{Results}
\label{Sec:3}
\begin{figure*}[ht]
	\begin{center}
		\includegraphics[width=\textwidth]{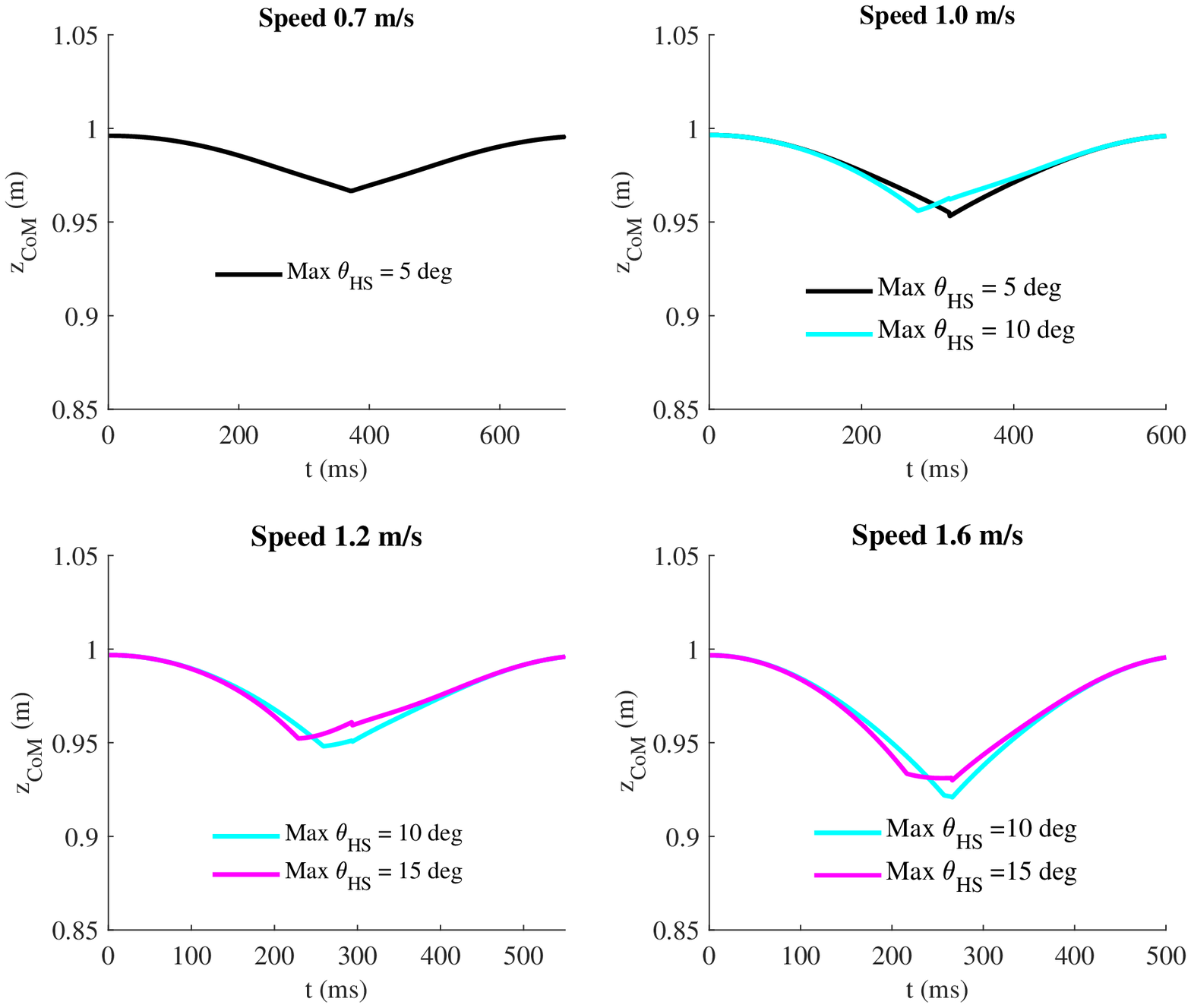}
	\end{center}
	\caption{The trajectories of the CoM vertical movements generated from the proposed planner show how the regulation of the ankle strategies during both Toe-Off and Hill-Strike can significantly alter the CoM trajectory. Although the CoM oscillation amplitude is much closer to human behaviour than traditional inverted pendulum models, an improvement in ankle strategies model is still required to obtain a cycloid shape observed in human trajectories \cite{Carpentier2017}.}
	\label{fig:09n}													
\end{figure*}

The results show that our model can reproduce the desired behaviour along the AP (Antero-Posterior) direction as shown in Figure \ref{fig:05}), while the behaviour along the ML (Medio-Lateral) direction as depicted Figure \ref{fig:07}, is within the human variability.Where it is shown that our planner can produce a behaviour which is compatible with human data from both the KITDB and Orendurff' s data \cite{Orendurff2004}. Furthermore, the vertical  trajectories (Figure \ref{fig:08}) are  sufficiently accurate for lower speeds, but they extend beyond human variability at the higher velocities. Proving how the ankle strategies allows humans to increase both the efficiency and the stability of their locomotion. Specifically, the increased length enables the CoM to reduce the inclination of the leg required to follow the desired trajectory on the transverse plane, thus minimizing the muscular effort via a more efficient redirection of the gravitational force through skeleton. Nevertheless, the error in tracking the vertical displacement of the CoM is significantly lower than the results obtained with a rigid pendulum. Although, the ankle strategies selected significantly improve the CoM vertical trajectory as shown in Figure \ref{fig:09n}, they are still not as smooth as the human cycloidal trajectories. 

\begin{figure*}
     \centering
     \begin{subfigure}[b]{0.49\textwidth}
         \centering
         \includegraphics[width=\textwidth]{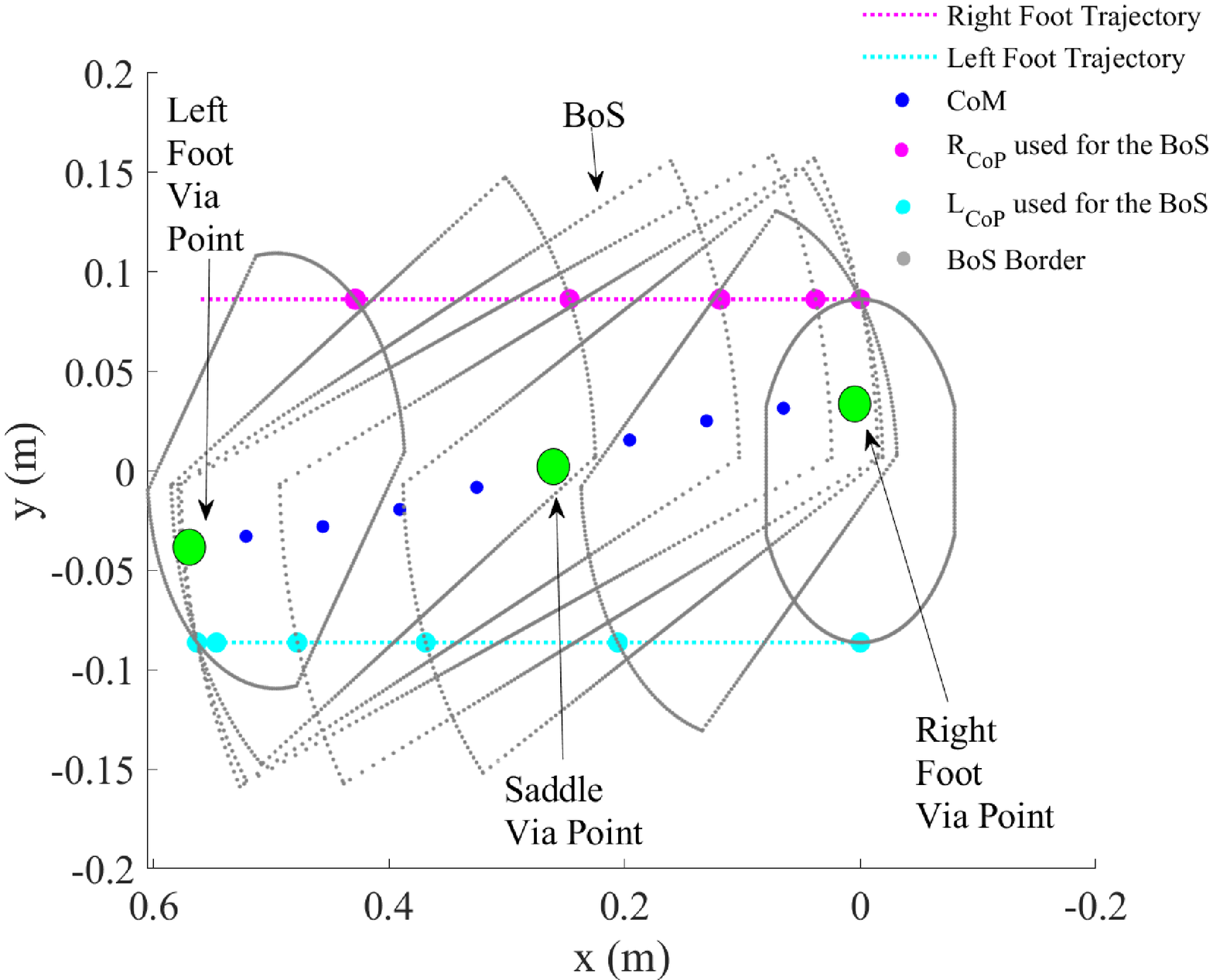}
         \caption{\centering}
         
         \label{fig:09}
     \end{subfigure}
     \begin{subfigure}[b]{0.49\textwidth}
         \centering
         \includegraphics[width=\textwidth]{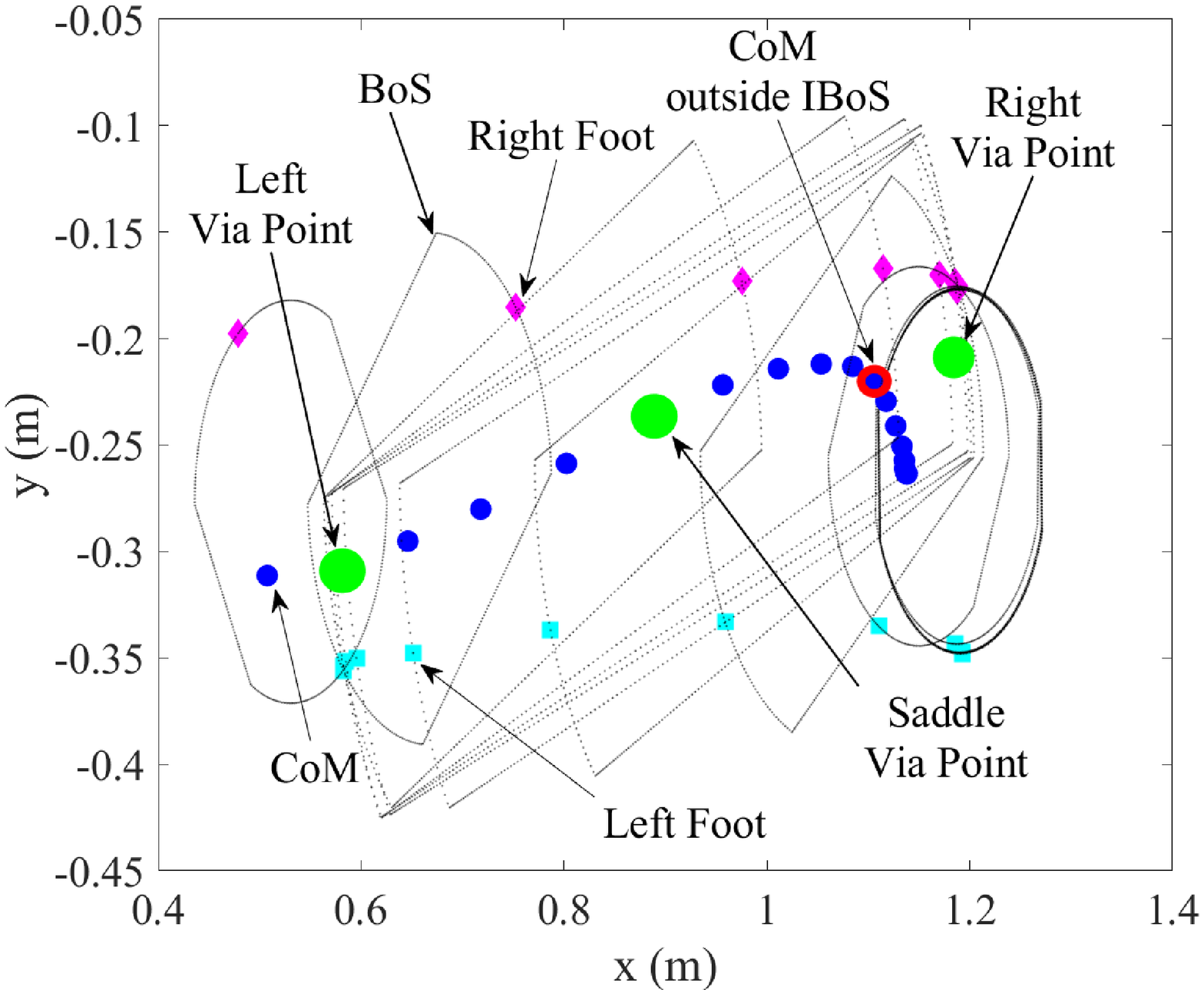}
         \caption{\centering}
         \label{fig:10}
     \end{subfigure}
        \caption{(a) The planner output between the right and left foot via points shows how the movement strategy allows to generate a funnel of stable points between the two legs. The generation of such funnel is necessary for the existence of a stable trajectory leading from one point to the other. (b) A human trajectory from the KITDB is shown to exemplify how the CoM moves towards a trajectory that is consistent with our planner before approaching the saddle via point. The analysis of KITDB data also suggests how humans may desynchronise the swinging leg to deploy the gravitational force to accelerate and decelerate the CoM. In conclusion, the comparison between the output of the proposed planner (a) and a human trajectory (b) supports our claim that the proposed architecture can produce human-like task-space planning for straight walking trajectories. }
       
\end{figure*}

Furthermore, the comparison between a trajectory from the proposed planner (Figure \ref{fig:09}) and a human trajectory from KITDB (Figure \ref{fig:10}) shows that human planning is consistent with our planner output. The human data also shows that the BoS tracks the movements of the CoM allowing the trajectory to occur on a funnel of points of equilibrium, which is a necessary condition for the Lyapunov's Stability. Lastly, there is a small portion of the trajectory which occurs outside of the BoS. Nonetheless, it does not compromise with the global stability of the system, because the swinging leg can reconfigure the system fast enough to drive the state back into the region of attraction before the foot landing occurs. These results suggest that the proposed model captures human planning strategies for both straight walking. 

\section{Discussion}
\label{Sec:4}
The obtained results indicate that it is possible to produce a computationally inexpensive human-like planning for straight walking using the Saddle Space model proposed in \cite{Tiseo2018a}. They also support our hypothesis that the divergence of the inverted pendulum model from the human behaviour can be justified by the presence of the ankle strategies. Although a simplified model used for the TO and HS movements does not provide an accurate human-like vertical trajectory, it can still generate a potential energy variation coherent with human behaviour, as shown in Figure \ref{fig:08}. 
\subsection{Proposed Stability Metrics}
The proposed stability metrics in this manuscript prove that is possible to evaluate the gait stability from the movement kinematics. Equation (\ref{eq:10}) combines the standard metrics in the kinematic constraints determined by the maximum reach of the leg with the dynamical constraints of the inverted pendulum models. Figure \ref{fig:05} shows how the planned strategy, that it is perfectly superimposed on the human reference behavior, is constrained from these conditions. Therefore, we can confirm the possibility of achieving a stable bipedal walking with a sufficiently long step length where the step frequency (candence) is greater than the natural frequency of inverted pendulum and smaller than the maximum reach of the leg. However, the legged locomotion is still achievable beyond the maximum reach of the legs using different strategies (i.e., running and jumping ). The other strategy described in equation (\ref{eq:11}) uses the step width and lateral excursion of the CoM to evaluate the mediolateral bipedal stability. Figure \ref{fig:08} elucidates that the human data from the two different datasets show the same trend of a slightly higher stability margin at higher speed. The numerical values obtained with the proposed planner are congruent with the range of variability observed in both data sets. 


\subsection{Considerations on the results generality}
Although our results are limited to the straight walking, the following general observation on to human locomotion can be ascertained:   
\begin{itemize}
	\item Human legs movement tends to synchronize with the CoM trajectory that shifts the CoM close to  $y_{Saddle}$. TSimilarly, the step to step transition tends to be along the direction of ZMP that simultaneously controls the angular momentums. \cite{Popovic2004,Popovic2004a,Pratt2006,Vlutters2016,Pratt2006}. Furthermore, the alignment of gravitational forces with $y_{Saddle}$ maximises the efficiency of the movement. 
	\item Adequate planning and control of the ankle strategies drastically improves  both the stability and efficiency of walking because they have a significant effect on the system energy expenditure. This observation is also supported by the other studies \cite{Kim2015,Ahn2012,Maki2007,Buschmann2015,Torricelli2016}.
	\item Our results also suggest that humans plan locomotion based on the predetermined optimised strategies. This is in agreement with the current motor control theories based on the observations that human movements are generated from a collection of stereotyped movements called dynamic primitives, which are influenced by external attractors \cite{Hogan2012,Zelik2014,Lacquaniti2012,Ajemian2010}. Therefore, the identification of simple kinematics parameters that allow reproducing a human-like walking suggests that locomotion can be regarded as a reaching task which is based on the dynamics constraints. 
\end{itemize}

\subsection{Integration in the Hierarchical Control Architecture}
\label{Sec:4.1}
Human motor-control seems to rely on a hierarchical control architecture. The higher modules take care of complex action planning in the task space, and operate at low frequencies. While descending the hierarchical structure, we encounter faster modules which plan and control less complex actions accordingly to the directives of the higher controllers \cite{Ahn2012,Brock2000}. The proposed method provides a novel approach to the higher level planning for human locomotion, which was not feasible with the previous models. Instead, our model allows to reduce the space of solution for the joint space planner via the introduction of constraints derived from the characterization of the gravitational force field. In other words, the proposed analytical model allows us to identify the desired foot movement for a given CoM trajectory or \textit{vice versa}. This imposes constraints in the joint space and limits the number of available solutions. 

The identification of the via points also potentially allows the integration of obstacle avoidance with the step planning via a modified version of elastic bands method proposed for autonomous robot navigation \cite{Quinlan1993,Brock2000}. Such approach should produce a more human-like behaviour in navigation, where humans often diverge from the theoretically optimal trajectory \cite{Sreenivasa2015}. Figure \ref{fig:08} also implies that the energy cost of a step is greatly affected from an inefficient ankle strategy. This leads us to hypothesise that it is better to implement a suboptimal navigation trajectory towards the final desired posture rather than computing a highly optimised navigation, which may also be invalidated by unforeseeable changes in the environment. For example, it has been recently observed how humans plan their navigation considering the final position of orientation. Therefore, they can converge gradually towards the desired trajectory without having a disrupting effect on their locomotion strategies \cite{Sreenivasa2015}. 

This theory is also supported by clinical data, which show the underlying relationship between balance and locomotion \cite{haruyama2017,Hof2007, shen2016, Lugade2011, hornby2016}. However, previous models identified balance as a consequence of the physiological gait parameters rather then the cause. Therefore, reconsidering the rehabilitation approach under this new insight may result in a better activation of the brain neuroplasticity.  

\subsection{Limitations and Future Developments}
The proposed planner is the first step in the development of a task-space planner for bipedal locomotion by taking the advantage of the intrinsic dynamics of the bipedal structure. However, the current results are limited to a simple task, and they do not take into account either for joint planning or multiple locomotion strategies. Thus further investigation is required to confirm our preliminary results and to validate the extendibility of this approach to different strategies.  In the next phase, we will integrate our proposed model with a joint planner as proposed in \cite{Tommasino2016,Tommasino2017} and evaluate the performance. 

\section{Conclusion}
\label{Sec:5}
The proposed bioinspired task-space planner for straight walking has been proven to produce accurate human-like trajectories. This study allow us to identify that the low variability behaviours (e.g. Step Length and CoM Transversal Trajectory) are mainly driven by the task-space planning, while the high variability is regulated by the lower controller to manage the local stability and optimisation (e.g. Step Width and vertical CoM trajectory). Furthermore, it also defines and validates methods for the stability supervision and analysis that relies only on the kinematics states. Furthermore, they allow us to explain how we can estimate the balance of other human beings from they locomotion kinematics. 

\section*{Acknowledgements}{This work is extracted from the PhD Thesis of Carlo Tiseo \cite{Tiseo2018a}. The authors would like to thank Mr Michele Xiloyannis and Dr Wouter Wolfslag for proofreading and reviewing the paper.This research was supported by the A*STAR-NHG-NTU Rehabilitation Research Grant: "Mobile Robotic Assistive Balance Trainer" (RRG/16018). The research work of Kalyana C. Veluvolu was supported by the National Research Foundation (NRF) of Korea funded by the Ministry of Education, Science and Technology under Grants (NRF-2017R1A2B2006032) and (NRF-2018R1A6A1A03025109).}

\section*{Conflicts of Interest}{The authors declare no conflict of interest.}


\bibliography{library}

\begin{thebibliography}{10}

\bibitem{Ahn2012}
J.~Ahn and N.~Hogan.
\newblock {Walking Is Not Like Reaching: Evidence from Periodic Mechanical
  Perturbations}.
\newblock {\em PLoS ONE}, 7(3):e31767, Mar 2012.

\bibitem{Ajemian2010}
R.~Ajemian and N.~Hogan.
\newblock {Experimenting with Theoretical Motor Neuroscience}.
\newblock {\em Journal of Motor Behavior}, 42(6):333--342, Oct 2010.

\bibitem{Brock2000}
O.~Brock and O.~Khatib.
\newblock {Real-time re-planning in high-dimensional configuration spaces using
  sets of homotopic paths}.
\newblock In {\em Proceedings 2000 ICRA. Millennium Conference. IEEE
  International Conference on Robotics and Automation. Symposia Proceedings
  (Cat. No.00CH37065)}, pages 550--555. IEEE, 2000.

\bibitem{Buschmann2015}
T.~Buschmann, A.~Ewald, A.~von Twickel, and A.~B{\"{u}}schges.
\newblock {Controlling legs for locomotion—insights from robotics and
  neurobiology}.
\newblock {\em Bioinspiration {\&} Biomimetics}, 10(4):041001, Jun 2015.

\bibitem{Caron2016}
S.~Caron, Q.-C. Pham, and Y.~Nakamura.
\newblock {ZMP Support Areas for Multicontact Mobility Under Frictional
  Constraints}.
\newblock {\em IEEE Transactions on Robotics}, pages 1--14, 2016.

\bibitem{Carpentier2017}
J.~Carpentier, M.~Benallegue, and J.-P. Laumond.
\newblock {On the centre of mass motion in human walking}.
\newblock {\em International Journal of Automation and Computing}, 2017.

\bibitem{Carpentier2016}
J.~Carpentier, S.~Tonneau, M.~Naveau, O.~Stasse, and N.~Mansard.
\newblock {A versatile and efficient pattern generator for generalized legged
  locomotion}.
\newblock In {\em 2016 IEEE International Conference on Robotics and Automation
  (ICRA)}, pages 3555--3561. IEEE, May 2016.

\bibitem{Collins2013}
S.~H. Collins and A.~D. Kuo.
\newblock {Two Independent Contributions to Step Variability during Over-Ground
  Human Walking}.
\newblock {\em PLoS ONE}, 8(8):e73597, Aug 2013.

\bibitem{dobkin2017}
B.~H. Dobkin.
\newblock A rehabilitation-internet-of-things in the home to augment motor
  skills and exercise training.
\newblock {\em Neurorehabilitation and neural repair}, 31(3):217--227, 2017.

\bibitem{englsberger2015three}
J.~Englsberger, C.~Ott, and A.~Albu-Sch{\"a}ffer.
\newblock Three-dimensional bipedal walking control based on divergent
  component of motion.
\newblock {\em IEEE Transactions on Robotics}, 31(2):355--368, 2015.

\bibitem{Farris2015}
D.~Farris, A.~Hampton, M.~D. Lewek, and G.~S. Sawicki.
\newblock {Revisiting the mechanics and energetics of walking in individuals
  with chronic hemiparesis following stroke: from individual limbs to lower
  limb joints}.
\newblock {\em Journal of NeuroEngineering and Rehabilitation}, 12(1):24, 2015.

\bibitem{Font-Llagunes2009}
J.~M. Font-Llagunes and J.~K{\"{o}}vecses.
\newblock {Dynamics and energetics of a class of bipedal walking systems}.
\newblock {\em Mechanism and Machine Theory}, 44(11):1999--2019, Nov 2009.

\bibitem{haruyama2017}
K.~Haruyama, M.~Kawakami, and T.~Otsuka.
\newblock Effect of core stability training on trunk function, standing
  balance, and mobility in stroke patients: a randomized controlled trial.
\newblock {\em Neurorehabilitation and neural repair}, 31(3):240--249, 2017.

\bibitem{Hof2005}
A.~Hof, M.~Gazendam, and W.~Sinke.
\newblock {The condition for dynamic stability}.
\newblock {\em Journal of Biomechanics}, 38(1):1--8, Jan 2005.

\bibitem{Hof2008}
A.~L. Hof.
\newblock {The ‘extrapolated center of mass' concept suggests a simple
  control of balance in walking}.
\newblock {\em Human Movement Science}, 27(1):112--125, Feb 2008.

\bibitem{Hof2007}
A.~L. Hof, R.~M. van Bockel, T.~Schoppen, and K.~Postema.
\newblock {Control of lateral balance in walking}.
\newblock {\em Gait {\&} Posture}, 25(2):250--258, Feb 2007.

\bibitem{Hogan2012}
N.~Hogan and D.~Sternad.
\newblock {Dynamic primitives of motor behavior}.
\newblock {\em Biological Cybernetics}, 106(11-12):727--739, Dec 2012.

\bibitem{hornby2016}
T.~G. Hornby, C.~L. Holleran, P.~W. Hennessy, A.~L. Leddy, M.~Connolly,
  J.~Camardo, J.~Woodward, G.~Mahtani, L.~Lovell, and E.~J. Roth.
\newblock Variable intensive early walking poststroke (views) a randomized
  controlled trial.
\newblock {\em Neurorehabilitation and neural repair}, 30(5):440--450, 2016.

\bibitem{Kim2015}
M.~Kim and S.~H. Collins.
\newblock {Once-per-step control of ankle-foot prosthesis push-off work reduces
  effort associated with balance during walking}.
\newblock {\em Journal of NeuroEngineering and Rehabilitation}, 12(1):43, Dec
  2015.

\bibitem{Kuo2007}
A.~D. Kuo.
\newblock {The six determinants of gait and the inverted pendulum analogy: A
  dynamic walking perspective}.
\newblock {\em Human Movement Science}, 26(4):617--656, Aug 2007.

\bibitem{Lacquaniti2012}
F.~Lacquaniti, Y.~P. Ivanenko, and M.~Zago.
\newblock {Patterned control of human locomotion}.
\newblock {\em The Journal of Physiology}, 590(10):2189--2199, May 2012.

\bibitem{Laumond2015}
J.-P. Laumond, N.~Mansard, and J.~B. Lasserre.
\newblock {Optimization as motion selection principle in robot action}.
\newblock {\em Communications of the ACM}, 58(5):64--74, Apr 2015.

\bibitem{Lei2006}
R.~Lei, H.~David, and K.~Laurence.
\newblock {Computational Models to Synthesize Human Walking}.
\newblock {\em Journal of Bionic Engineering}, 3(3):127--138, Sep 2006.

\bibitem{Lugade2011}
V.~Lugade, V.~Lin, and L.-s. Chou.
\newblock {Center of mass and base of support interaction during gait}.
\newblock {\em Gait {\&} Posture}, 33(3):406--411, Mar 2011.

\bibitem{Majumdar2013}
A.~Majumdar and R.~Tedrake.
\newblock {Robust Online Motion Planning with Regions of Finite Time
  Invariance}.
\newblock In E.~Frazzoli, T.~Lozano-Perez, N.~Roy, and D.~Rus, editors, {\em
  Algorithmic Foundations of Robotics {\ldots}}, Springer Tracts in Advanced
  Robotics, pages 543--558. Springer Berlin Heidelberg, Berlin, Heidelberg,
  2013.

\bibitem{Maki2007}
B.~E. Maki and W.~E. McIlroy.
\newblock {Cognitive demands and cortical control of human balance-recovery
  reactions}.
\newblock {\em Journal of Neural Transmission}, 114(10):1279--1296, Oct 2007.

\bibitem{Manchester2010}
I.~R. Manchester, M.~M. Tobenkin, M.~Levashov, and R.~Tedrake.
\newblock {Regions of Attraction for Hybrid Limit Cycles of Walking Robots}.
\newblock {\em IFAC Proceedings Volumes}, 44(1):5801----5806, Oct 2010.

\bibitem{Mandery2015}
C.~Mandery, O.~Terlemez, M.~Do, N.~Vahrenkamp, and T.~Asfour.
\newblock {The KIT whole-body human motion database}.
\newblock In {\em 2015 International Conference on Advanced Robotics (ICAR)},
  pages 329--336. IEEE, jul 2015.

\bibitem{McAndrewYoung2012}
P.~M. {McAndrew Young} and J.~B. Dingwell.
\newblock {Voluntary changes in step width and step length during human walking
  affect dynamic margins of stability}.
\newblock {\em Gait {\&} Posture}, 36(2):219--224, Jun 2012.

\bibitem{McGeer1990}
T.~McGeer.
\newblock {Passive Dynamic Walking}.
\newblock {\em The International Journal of Robotics Research}, 9(2):62--82,
  Apr 1990.

\bibitem{McIlroy2003}
W.~E. McIlroy, D.~C. Bishop, W.~R. Staines, A.~J. Nelson, B.~E. Maki, and J.~D.
  Brooke.
\newblock {Modulation of afferent inflow during the control of balancing tasks
  using the lower limbs.}
\newblock {\em Brain research}, 961(1):73--80, Jan 2003.

\bibitem{MyungheeKim2013}
{Myunghee Kim} and S.~H. Collins.
\newblock {Stabilization of a three-dimensional limit cycle walking model
  through step-to-step ankle control}.
\newblock In {\em 2013 IEEE 13th International Conference on Rehabilitation
  Robotics (ICORR)}, pages 1--6. IEEE, Jun 2013.

\bibitem{Orendurff2004}
M.~S. Orendurff, A.~D. Segal, G.~K. Klute, J.~S. Berge, E.~S. Rohr, and N.~J.
  Kadel.
\newblock {The effect of walking speed on center of mass displacement.}
\newblock {\em Journal of rehabilitation research and development},
  41(6A):829--34, 2004.

\bibitem{Patnaik2015}
L.~Patnaik and L.~Umanand.
\newblock {Physical constraints, fundamental limits, and optimal locus of
  operating points for an inverted pendulum based actuated dynamic walker}.
\newblock {\em Bioinspiration and Biomimetics}, 10(6), 2015.

\bibitem{Perrin2012}
N.~Perrin, O.~Stasse, L.~Baudouin, F.~Lamiraux, and E.~Yoshida.
\newblock {Fast Humanoid Robot Collision-Free Footstep Planning Using Swept
  Volume Approximations}.
\newblock {\em IEEE Transactions on Robotics}, 28(2):427--439, Apr 2012.

\bibitem{Popovic2004}
M.~Popovic, A.~Hofmann, and H.~Herr.
\newblock {Angular momentum regulation during human walking: biomechanics and
  control}.
\newblock In {\em IEEE International Conference on Robotics and Automation,
  2004. Proceedings. ICRA '04. 2004}, pages 2405--2411 Vol.3. IEEE, 2004.

\bibitem{Popovic2004a}
M.~Popovic, A.~Hofmann, and H.~Herr.
\newblock {Zero spin angular momentum control: definition and applicability}.
\newblock In {\em 4th IEEE/RAS International Conference on Humanoid Robots,
  2004.}, volume~1, pages 478--493. IEEE, 2004.

\bibitem{Pratt2006}
J.~Pratt and R.~Tedrake.
\newblock {Velocity-Based Stability Margins for Fast Bipedal Walking}.
\newblock In {\em Fast Motions in Biomechanics and Robotics}, pages 299--324.
  Springer Berlin Heidelberg, Berlin, Heidelberg, 2006.

\bibitem{Quinlan1993}
S.~Quinlan and O.~Khatib.
\newblock {Elastic bands: connecting path planning and control}.
\newblock In {\em [1993] Proceedings IEEE International Conference on Robotics
  and Automation}, pages 802--807. IEEE Comput. Soc. Press, 1993.

\bibitem{SAUNDERS1953}
J.~B. Saunders, V.~T. Inman, and H.~D. Eberhart.
\newblock {The major determinants in normal and pathological gait.}
\newblock {\em The Journal of bone and joint surgery. American volume},
  35-A(3):543--58, jul 1953.

\bibitem{shen2016}
X.~Shen, I.~S. Wong-Yu, and M.~K. Mak.
\newblock Effects of exercise on falls, balance, and gait ability in
  parkinson’s disease: a meta-analysis.
\newblock {\em Neurorehabilitation and neural repair}, 30(6):512--527, 2016.

\bibitem{Sreenivasa2015}
M.~Sreenivasa, K.~Mombaur, and J.-P. Laumond.
\newblock {Walking Paths to and from a Goal Differ: On the Role of Bearing
  Angle in the Formation of Human Locomotion Paths}.
\newblock {\em PLOS ONE}, 10(4):e0121714, Apr 2015.

\bibitem{Tiseo2018a}
C.~Tiseo.
\newblock {\em {Modelling of bipedal locomotion for the development of a
  compliant pelvic interface between human and a balance assistant robot}}.
\newblock PhD thesis, Nanyang Technological University, 2018.

\bibitem{Tiseo2016}
C.~Tiseo and W.~Ang.
\newblock {The Balance: An energy management task}.
\newblock In {\em Proceedings of the IEEE RAS and EMBS International Conference
  on Biomedical Robotics and Biomechatronics}, volume 2016-July, pages
  723--728, 2016.

\bibitem{Tiseo2018c}
C.~Tiseo, M.~J. Foo, K.~C. Veluvolu, and A.~W. Tech.
\newblock {A Postural Model for Tracking the Base of Support}.
\newblock In {\em 2018 40th Annual International Conference of the IEEE
  Engineering in Medicine and Biology Society, EMBC 2018}, pages 1833--1836,
  2018.

\bibitem{Tiseo2018b}
C.~Tiseo, K.~C. Veluvolu, and W.~T. Ang.
\newblock {Evidence of a “ Clock ” Determining Human Locomotion}.
\newblock In {\em 2018 40th Annual International Conference of the IEEE
  Engineering in Medicine and Biology Society, EMBC 2018}, pages 1693--1696,
  2018.

\bibitem{Tiseo2018d}
C.~Tiseo, K.~C. Veluvolu, and W.~T. Ang.
\newblock {The bipedal Saddle Space: Modelling and validation}.
\newblock {\em Bioinspiration {\&} Biomimetics}, pages 0--7, Oct 2018.

\bibitem{Tommasino2016}
P.~Tommasino and D.~Campolo.
\newblock {Human-like pointing strategies via non-linear inverse optimization}.
\newblock In {\em 2016 6th IEEE International Conference on Biomedical Robotics
  and Biomechatronics (BioRob)}, volume 2016-July, pages 930--935. IEEE, Jun
  2016.

\bibitem{Tommasino2017}
P.~Tommasino and D.~Campolo.
\newblock {Task-space separation principle: a force-field approach to motion
  planning for redundant manipulators}.
\newblock {\em Bioinspiration {\&} Biomimetics}, 12(2):026003, Feb 2017.

\bibitem{Torricelli2016}
D.~Torricelli, J.~Gonzalez, M.~Weckx, R.~Jim{\'{e}}nez-Fabi{\'{a}}n,
  B.~Vanderborght, M.~Sartori, S.~Dosen, D.~Farina, D.~Lefeber, and J.~L. Pons.
\newblock {Human-like compliant locomotion: state of the art of robotic
  implementations}.
\newblock {\em Bioinspiration {\&} Biomimetics}, 11(5):051002, Aug 2016.

\bibitem{VanMeulen2016}
F.~B. van Meulen, D.~Weenk, E.~H.~F. van Asseldonk, H.~M. Schepers, P.~H.
  Veltink, and J.~H. Buurke.
\newblock {Analysis of Balance during Functional Walking in Stroke Survivors}.
\newblock {\em PLOS ONE}, 11(11):e0166789, Nov 2016.

\bibitem{Virmavirta2014}
M.~Virmavirta and J.~Isolehto.
\newblock {Determining the location of the body s center of mass for different
  groups of physically active people}.
\newblock {\em Journal of Biomechanics}, 47(8):1909--1913, Jun 2014.

\bibitem{Vlutters2016}
M.~Vlutters, E.~H.~F. van Asseldonk, and H.~van~der Kooij.
\newblock {Center of mass velocity-based predictions in balance recovery
  following pelvis perturbations during human walking}.
\newblock {\em The Journal of Experimental Biology}, 219(10):1514--1523, May
  2016.

\bibitem{Zelik2014}
K.~E. Zelik, V.~{La Scaleia}, Y.~P. Ivanenko, and F.~Lacquaniti.
\newblock {Can modular strategies simplify neural control of multidirectional
  human locomotion?}
\newblock {\em Journal of Neurophysiology}, 111(8):1686--1702, Apr 2014.

\end{thebibliography}

\bibliographystyle{abbrv}

\end{document}